\documentclass[sigconf]{acmart}
\usepackage{bbm}
\usepackage{amsmath}
\usepackage{multirow}
\usepackage{makecell}

\AtBeginDocument{%
  \providecommand\BibTeX{{%
    \normalfont B\kern-0.5em{\scshape i\kern-0.25em b}\kern-0.8em\TeX}}}

\setcopyright{rightsretained}
\copyrightyear{2018}
\acmYear{2018}
\acmDOI{10.1145/1122445.1122456}

\acmConference[KDD '21]{KDD '21:  Conference on Knowledge Discovery and Data Mining }{August 14--18, 2021}{Virtual Event, Singapore}
\acmBooktitle{KDD '21:  Conference on Knowledge Discovery and Data Mining,
  August 14--18, 2021, Virtual Event, Singapore}
\acmPrice{15.00}
\acmISBN{978-1-4503-XXXX-X/18/06}

\begin{document}

\title{PcDGAN: A Continuous Conditional Diverse Generative Adversarial Network For Inverse Design}

\author{Amin Heyrani Nobari}
\affiliation{%
  \institution{Massachusetts Institute of Technology}
  \city{Cambridge, MA}
  \country{USA}}
\email{ahnobari@mit.edu}

\author{Wei Chen}
\affiliation{%
  \institution{Siemens Corporate Technology}
  \city{Princeton, NJ}
  \country{USA}}
  \email{chen.wei@siemens.com}

\author{Faez Ahmed}
\affiliation{%
 \institution{Massachusetts Institute of Technology}
  \city{Cambridge, MA}
  \country{USA}}
  \email{faez@mit.edu}
\copyrightyear{2021}
\acmYear{2021}
\acmConference[KDD '21]{Proceedings of the 27th ACM SIGKDD Conference on Knowledge Discovery and Data Mining}{August 14--18, 2021}{Virtual Event, Singapore}
\acmBooktitle{Proceedings of the 27th ACM SIGKDD Conference on Knowledge Discovery and Data Mining (KDD '21), August 14--18, 2021, Virtual Event, Singapore}\acmDOI{10.1145/3447548.3467414}
\acmISBN{978-1-4503-8332-5/21/08}

\newcommand{\eg}{{\em e.g.}}
\newcommand{\etal}{{\em et~al.}}
\newcommand{\ie}{{\em i.e.}}
\newcommand{\etc}{{\em etc.}}
\newcommand{\RNum}[1]{\uppercase\expandafter{\romannumeral #1\relax}}

\begin{abstract}
  Engineering design tasks often require synthesizing new designs that meet desired performance requirements. The conventional design process, which requires iterative optimization and performance evaluation, is slow and dependent on initial designs. Past work has used conditional generative adversarial networks (cGANs) to enable direct design synthesis for given target performances. However, most existing cGANs are restricted to categorical conditions. Recent work on Continuous conditional GAN (CcGAN) tries to address this problem, but still faces two challenges: 1) it performs poorly on non-uniform performance distributions, and 2) the generated designs may not cover the entire design space. We propose a new model, named Performance Conditioned Diverse Generative Adversarial Network (PcDGAN), which introduces a singular vicinal loss combined with a Determinantal Point Processes (DPP) based loss function to enhance diversity. PcDGAN uses a new self-reinforcing score called the Lambert Log Exponential Transition Score (LLETS) for improved conditioning. Experiments on synthetic problems and a real-world airfoil design problem demonstrate that PcDGAN outperforms state-of-the-art GAN models and improves the conditioning likelihood by 69\% in an airfoil generation task and upto 78\% in synthetic conditional generation tasks and achieves greater design space coverage. The proposed method enables efficient design synthesis and design space exploration with applications ranging from CAD model generation to metamaterial selection.
\end{abstract}


\maketitle

\section{Introduction}
Engineering design applications often aim to synthesize a set of designs that are diverse and meet specific performance requirements.
Designers, therefore, regularly spend significant time exploring the \textit{design space} (\ie, the space of all possible design options) to find desired solutions. In the process, they have to repetitively evaluate numerous options, often using physics-based simulation models, and make adjustments depending on how a design meets each requirement. The engineering design community has made huge efforts in accelerating or eliminating this tedious trial-and-error design cycle. For example, topology optimization~\cite{duysinx1998topology,bendsoe2013topology} and adjoint-based optimization~\cite{anderson1999aerodynamic} are commonly used in structural design and aerodynamic design, respectively. These methods exclude humans from the design cycle and can automatically generate solutions based on performance requirements. However, they are still time-consuming due to expensive design evaluations (\ie, usually through finite element analysis or computational fluid dynamics simulation). Also, those methods require differentiable physics solvers so that gradients can be evaluated for each design option in the design optimization process. This makes it impossible to apply those methods to scenarios where performance is evaluated by non-analytical models (\eg, experiments or expert assessments). This brings up a question: can we skip the trial-and-error process or the expensive simulations and obtain a design that satisfy any given performance requirements? This is a so-called \textit{inverse design} problem.

To solve this problem, recent work has also looked at data-driven methods such as conditional deep generative models~\cite{yilmaz2020conditional,achour2020development}.
Deep generative models like generative adversarial networks (GANs)~\cite{GAN} and variational autoencoders (VAEs)~\cite{VAE} are often good at learning the complex distribution of existing designs. This allows designers to quickly synthesize new designs by sampling from that distribution. By using models like conditional GANs (cGANs)~\cite{CGAN} or conditional VAEs (CVAEs)~\cite{sohn2015learning}, it is possible to condition the synthesized designs on arbitrary performance requirements, and thus completely eliminate the tedious trial-and-error process and the expensive simulations.

Recent developments in conditional deep generative models have proven successful in many different generative tasks like image generation, where the conditions are usually categorical class labels. This assumption is impractical in many engineering design settings since performance is usually continuous.
For example, in turbine design, some important performance metrics are the power coefficient, the pressure coefficient, and the cavitation number~\cite{TurbineExample}; in aerodynamic design, performance is measured by lift to drag ratio or inverse lift coefficient~\cite{BezierGAN}; in beam design, common metrics are compliance and natural-frequency~\cite{BeamExamples} \textemdash these metrics are all continuous variables. To use those metrics as the condition in conditional generative models, past work~\cite{yilmaz2020conditional,achour2020development} proposes to discretize the continuous values of the metrics to discrete bins. However, this approximation leads to poor learning, as the labels lose order information given their values will be rounded to the nearest bin.
To overcome these limitations, researchers have recently proposed methods for conditioning GANs in continuous spaces. The state-of-the-art method is the continuous conditional GAN (CcGAN)~\cite{CcGAN}. However, though it demonstrates good performance on computer vision applications, it is not sufficient for engineering settings, where 1)~the condition satisfaction is much more strict and 2)~the diversity of synthesized designs is highly encouraged.

In this paper, we propose a new GAN model, named Performance conditioned Diverse Generative Adversarial Network (PcDGAN), which has two main objectives: 1)~to generate design candidates that have a good coverage of the entire design space, \ie, to maximize the diversity of the generated samples, and 2)~to generate design candidates which meet any given scalar performance requirement. We achieve these objectives by combining a conditioning-driven determinantal point processes (DPP) loss to capture sample diversity and promote conditioning on top of a new loss term, which we name singular vicinal loss, to further capture the conditioning performance. 
We show that PcDGAN not only allows efficient design synthesis given continuous conditions (\ie, requirements on continuous performance metrics), but also ensures a good coverage of the design space, despite facing data sparsity problem.
PcDGAN is able to drastically improve the pace at which automated design space exploration for specific tasks can be accomplished. In this paper, we describe the key innovations in PcDGAN and show that PcDGAN is capable of outperforming the current state-of-the-art in continuous conditioning of GANs (\ie, CcGAN) in design synthesis applications. We summarize main contributions as follows:
\begin{enumerate}
\item We propose a novel discriminator loss, the singular vicinal loss, and combine it with a DPP loss to promote conditional diversity. We demonstrate that the proposed loss improves the diversity score by an average of 14\% compared to CcGAN in an airfoil generation task and 21\% in a synthetic imbalanced conditional generation task.
\item We introduce a novel scoring function, the Lambert log exponential transition score (LLETS), and incorporate it into the DPP loss to ensure accurate conditioning. We demonstrate that this scoring function improves the conditioning likelihood score by 69\% in an airfoil generation task and upto 78\% in synthetic conditional generation tasks.
\item We propose the integration of regression estimators (neural networks-based or exact) into the process of conditioning in continuous spaces, which to the best of the authors' knowledge is the first model to do so.
\item We apply the proposed method on a real-world inverse airfoil design problem and show that PcDGAN overcomes the issue of severe imbalances in the data labels and outperforms state of art CcGAN method on diversity and conditioning metrics, while also having lower variation across different runs.
\end{enumerate}

\section{Background and Related Work}
In this section, we provide a concise background on two topics explored in this work: conditional generative adversarial networks and determinantal point processes.

\subsection{Continuous Conditioning of Generative Adversarial Networks}
The conditional GAN (cGAN)~\cite{CGAN} learns the distribution of samples conditioned on some auxiliary information. Such information is usually the class labels of images, which are categorical variables. Ding~\etal~\cite{CcGAN} proposed the continuous conditional GAN (CcGAN) to address the challenges in cGAN when the conditions are continuous labels.
One challenge is that cGAN's empirical loss function relies on a large number of samples for each distinct condition as empirical risk minimization (ERM) approaches often do~\cite{vapnik_2000}. This makes it unsuitable for conditioning in continuous spaces where some labels may have few or even no samples associated with them. To address this, CcGAN uses a novel loss function for the discriminator called the \textit{vicinal loss}, which is based around the principles of vicinal risk minimization(VRM)~\cite{vapnik_2000,Chapelle01vicinalrisk}. They propose two variants of this loss function \textemdash the \textit{hard vicinal discriminator loss} (HVDL) and the \textit{soft vicinal discriminator loss} (SVDL). Rather than applying the cGAN loss over the entire data at every training step, the loss is applied to samples in the vicinity (in the condition space) of selected labels from the data:
\begin{equation}
\label{eqn:3}
\begin{split}
\resizebox{.91\linewidth}{!}{$\widehat{\mathcal{L}}^{\mathrm{HVDL}}(D)=\qquad\qquad\qquad\qquad\qquad\qquad\qquad\qquad\qquad\qquad\qquad\qquad$} \\\resizebox{.91\linewidth}{!}{$-\frac{C_{1}}{N^{r}} \sum_{j=1}^{N^{r}} \sum_{i=1}^{N^{r}} \mathbb{E}_{\epsilon^{r} \sim \mathcal{N}\left(0, \sigma^{2}\right)}\left[\frac{\mathbbm{1}_{\left\{\left|y_{j}^{r}+\epsilon^{r}-y_{i}^{r}\right| \leq \kappa\right\}}}{N_{y_{j}^{r}+\epsilon^{r}, \kappa}^{r}} \log \left(D\left(\boldsymbol{x}_{i}^{r}, y_{j}^{r}+\epsilon^{r}\right)\right)\right]$} \\
\resizebox{.91\linewidth}{!}{$-\frac{C_{2}}{N^{g}} \sum_{j=1}^{N^{g}} \sum_{i=1}^{N^{g}} \mathbb{E}_{\epsilon^{g} \sim \mathcal{N}\left(0, \sigma^{2}\right)}\left[\frac{\mathbbm{1}_{\left\{\left|y_{j}^{g}+\epsilon^{g}-y_{i}^{g}\right| \leq \kappa\right\}}}{N_{y_{j}^{g}+\epsilon^{g}, \kappa}^{g}} \log \left(1-D\left(\boldsymbol{x}_{i}^{g}, y_{j}^{g}+\epsilon^{g}\right)\right)\right]$},
\end{split}
\end{equation}

\begin{equation}
\label{eqn:4}
\begin{split}
\resizebox{.91\linewidth}{!}{$\widehat{\mathcal{L}}^{\mathrm{SVDL}}(D)=\qquad\qquad\qquad\qquad\qquad\qquad\qquad\qquad\qquad\qquad\qquad\qquad$} \\\resizebox{.91\linewidth}{!}{$-\frac{C_{3}}{N^{r}} \sum_{j=1}^{N^{r}} \sum_{i=1}^{N^{r}} \mathbb{E}_{\epsilon^{r} \sim \mathcal{N}\left(0, \sigma^{2}\right)}\left[\frac{w^{r}\left(y_{i}^{r}, y_{j}^{r}+\epsilon^{r}\right)}{\sum_{i=1}^{N^{r}} w^{r}\left(y_{i}^{r}, y_{j}^{r}+\epsilon^{r}\right)} \log \left(D\left(\boldsymbol{x}_{i}^{r}, y_{j}^{r}+\epsilon^{r}\right)\right)\right]$} \\
\resizebox{.91\linewidth}{!}{$-\frac{C_{4}}{N^{g}} \sum_{j=1}^{N^{g}} \sum_{i=1}^{N^{g}} \mathbb{E}_{\epsilon^{g} \sim \mathcal{N}\left(0, \sigma^{2}\right)}\left[\frac{w^{g}\left(y_{i}^{g}, y_{j}^{g}+\epsilon^{g}\right)}{\sum_{i=1}^{N^{g}} w^{g}\left(y_{i}^{g}, y_{j}^{g}+\epsilon^{g}\right)} \log \left(1-D\left(\boldsymbol{x}_{i}^{g}, y_{j}^{g}+\epsilon^{g}\right)\right)\right]$},
\end{split}
\end{equation}
where $y$ denotes the label of the sample $x$. Hyper-parameters $\sigma$ and $\kappa$ determine the width of the vicinity in conditioning space. $C_{1}$, $C_{2}$, $C_{3}$, and $C_{4}$ are constants. $N$ is the number of samples. Superscripts $r$ and $g$ denotes real and generated samples, respectively. For SVDL, weights are applied to the loss values and are computed as:
\begin{equation}
\label{eqn:5}
w^{r}\left(y_{i}^{r}, y\right)=e^{-\nu\left(y_{i}^{r}-y\right)^{2}} \text { and } w^{g}\left(y_{i}^{g}, y\right)=e^{-\nu\left(y_{i}^{g}-y\right)^{2}}.
\end{equation}
Essentially, this loss is calculated by selecting samples with labels $y$ uniformly sampled from the data, adding random noise $\epsilon$ to the selected labels $y$, and selecting samples such that their labels $y$ are in the vicinity of $y+\epsilon$. The same is done to obtain targets to generate fake samples by the generator. Given these samples, the discriminator is trained by minimizing Eq.\ref{eqn:3} or Eq.\ref{eqn:4}. The generator is then trained on targets in a similar vicinity of the samples the discriminator was trained on, by minimizing the following loss:
\begin{equation}
\label{eqn:6}
\begin{split}
\widehat{\mathcal{L}}^{\epsilon}(G)=\qquad\qquad\qquad\qquad\qquad\qquad\qquad\qquad\qquad\qquad\\
\resizebox{.91\linewidth}{!}{$-\frac{1}{N^{g}} \mathbb{E}_{\epsilon^{g} \sim \mathcal{N}\left(0, \sigma^{2}\right)} \log \left(D\left(G\left(\boldsymbol{z}_{i}, y_{i}^{g}+\epsilon^{g}\right), y_{i}^{g}+\epsilon^{g}\right)\right)$}.
\end{split}
\end{equation}
Another challenge of cGAN that Ding~\etal notes is the method of feeding conditions~\cite{CcGAN}. They point out that when conditions exist in finite space, one-hot encoding of the labels can be used, however, this is simply not practical in continuous spaces as finite distinct labels do not exit. They propose the improved label input mechanism (ILI). It trains an embedder model that maps the conditions (from data) to the features extracted from the last layer of a pre-trained regression model. The embedder model is then used during the GAN training to map arbitrary conditions to feature vectors, which are fed to the generator through conditional batch normalization~\cite{cbn}. The same embedded features are also used to condition the discriminator. 

We investigate the application of CcGAN in real-world and synthetic examples and observe that CcGAN does not perform well in conditioning especially when the label distribution is uneven with sparse distribution of labels in some parts of the label space. Further, we observe that CcGAN can suffer in diversity when the data distribution in the design space is imbalanced. In this paper, we propose a novel approach in conditioning in continuous spaces to improve GAN performance in these applications.

\subsection{Determinantal Point Processes}
Determinantal Point Processes (DPPs), which arise in quantum physics, are probabilistic models that model the likelihood of selecting a subset of diverse items as the determinant of a kernel matrix. Viewed as joint distributions over the binary variables corresponding to item selection, DPPs essentially capture negative correlations and provide a way to elegantly model the trade-off between often competing notions of quality and diversity. The intuition behind DPPs is that the determinant of a kernel matrix roughly corresponds to the volume spanned by the vectors representing the items. Points that “cover” the space well should capture a larger volume of the overall space, and thus have a higher probability. For the purpose of working with most types of data the DPP kernel can be constructed using the L-ensembles approach~\cite{lensemble}. An L-ensemble defines a DPP via a positive semi-definite matrix $L$ indexed by the elements of a subset $S$. The kernel matrix L defines a global measure of similarity between pairs of items, so that more similar items are less likely to co-occur. The probability of a set $S$ occurring under a DPP is calculated as:
\begin{equation}
\mathbb{P}_{L}(S)=\frac{\operatorname{det}\left(L_{S}\right)}{\operatorname{det}(L+I)},
\end{equation}
where $L_{s}=[L_{ij}]_{i,j \in S}$ denotes the restriction of L to the entries indexed by elements of $S$. DPP kernels can be decomposed into quality and diversity parts~\cite{Kulesza_2012}. The $(i, j)$-th entry of a positive semi-definite DPP kernel L can be expressed as:
\begin{equation}
\label{eqn:8}
L_{i j}=q_{i} \phi(i)^{T} \phi(j) q_{j},
\end{equation}
We can think of $q_{i} \in R^{+}$ as a scalar value measuring the quality of an item $i$, and $\phi(i)^T\phi(j)$ as a signed measure of similarity between items $i$ and $j$. The decomposition enforces $L$ to be positive semi-definite. Suppose we select a subset $S$ of samples, then this decomposition allows us to write the probability of this subset $S$ as the square of the volume spanned by $q_{i}\phi_{i}$ for $i\in S$ using the equation below:
\begin{equation}
\mathbb{P}_{L}(S) \propto \prod_{i \in S}\left(q_{i}^{2}\right) \operatorname{det}\left(K_{S}\right),
\end{equation}
where $K_{S}$ is the similarity matrix of $S$. As item $i$'s quality $q_{i}$ increases, so do the probabilities of sets containing item $i$. As two items $i$ and $j$ become more similar, $\phi(i)^T\phi(j)$ increases and the probabilities of sets containing both $i$ and $j$ decrease. The previous model, PaDGAN, introduces a DPP loss to maximize the DPP probability~\cite{padgan}. This simultaneously encourages a larger coverage of the data space and high-quality sample generation. In PcDGAN, we will also use this loss to promote good conditioning, which is elaborated in the next section.

\section{Methodology}

\begin{figure*}[ht!]
\centering
\includegraphics[width=1.8\columnwidth]{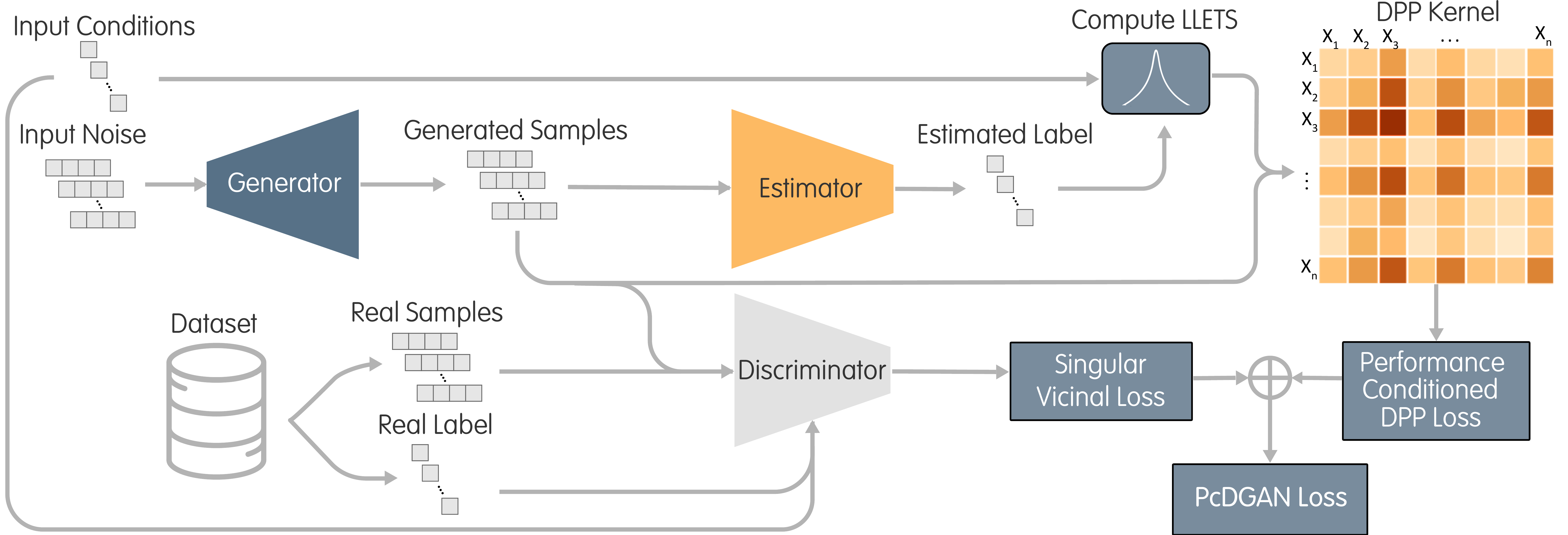}
\vskip -0.1in
\caption{PcDGAN architecture}
\label{fig:architecture}
\end{figure*}

In PcDGAN, we introduce an alternative perspective to conditioning GANs on continuous labels. The overall architecture of PcDGAN is shown in Fig.~\ref{fig:architecture}. Specifically, we approach the problem from an engineering design prospective, where the objective is to generate designs that (1)~cover the entire design space evenly, and (2)~meet specific performance requirements (\ie, conditions). To address these, we add a performance conditioned DPP loss and introduce a different variant of the vicinal loss called the \textit{singular vicinal loss}, which overcomes the issue of data imbalance. Furthermore, the singular vicinal loss promotes design space coverage for all points in the label (performance) space. We describe the details of each part in this section.

\subsection{Determining the Quality of Conditioning}
In cGAN and CcGAN, the discriminator must distinguish between real and generated samples by using both data and labels (conditions), effectively making it a vanilla discriminator and a label predictor simultaneously. 
In PcDGAN, to enable improved conditioning, we add a separate \textit{performance estimator}. The performance estimator is any differentiable estimator that can predict the label (\ie, design performance in our case) of any generated or real samples. In many engineering design applications, we have high fidelity, differentiable physics-based performance estimators (\eg, adjoint CFD solvers). When such a differentiable estimator does not exist, we can train a neural network-based regressor as the performance estimator.

PcDGAN enforces the conditioning by introducing a \textit{conditioning quality} term to the DPP loss. Specifically, the conditioning quality reflects how well the condition is met by the generated samples. It is used as the quality terms in the DPP kernel ($q_{i} \text{and} q_{j}$ in Eq.\ref{eqn:8}). A straightforward way to measure the conditioning quality could be using the negative norm (L1 or L2) of the difference between the estimator prediction and the desired condition. This approach in our experiments fails to provide good conditioning, possibly because of the small gradients of the L1 (which is either 1 or -1) and L2 norms, when the labels are normalized to range from 0 to 1. Furthermore, the gradient of the L1 error is always constant and the gradient of L2 error increases with poor conditioning which is not desirable when applying the resultant value as a (conditioning) quality term in the DPP kernel. 

Research at the intersection of reinforcement learning and GANs has shown that increasing the weight, hence gradients, of samples that have desirable features, reinforces the GANs to produce samples with such desirable features. This is demonstrated in objective-reinforced GAN (ORGAN)~\cite{organ}, where objectives are reinforced by weighing their gradients relative to their performance. However, ORGAN required determining the weights of samples.
In contrast, we introduce an approach to systematically do this without having to determine weights. We first list the properties needed in a conditioning quality metric: 1)~the quality score for each label needs to have the same scale to avoid biasing the DPP probability function; 2)~the gradient of the score function should ideally increase as the conditioning quality increases, such that the score becomes self reinforcing; 3)~the gradient should smooth out beyond a certain point to allow for stabilization at high quality; and 4)~the score must only have zero gradient at the origin (when the difference between estimator prediction and the desired label is zero corresponding to perfect conditioning). With these considerations, we introduce a scoring function called \textit{Lambert log exponential transition score} (LLETS) which is expressed as
\begin{equation}
\label{eqn:10}
LLETS(\epsilon)=\left\{\begin{array}{ll}
-\frac{\ln \epsilon}{a} & \epsilon>e^{-a e^{W\left(-\frac{1}{2 a}\right)}} \\
e^{-\frac{\epsilon^{2}}{2 \sigma}} & \epsilon \leq e^{-a e^{W\left(-\frac{1}{2 a}\right)}}
\end{array}\right.,
\end{equation}
where $\epsilon=\|y_{\text{condition}}-y_{\text{estimated}}\|_1$ is the L1 error between the estimator prediction and the desired label. We scale the labels between 0 and 1 so that $\epsilon$ is also between 0 and 1. The Lambert cutoff $a\geq e/2$ is a hyper-parameter determining the location and quantity of the maximum gradient. $W$ refers to the Lambert W function and $\sigma$ is determined by:
\begin{equation}
    \sigma=\frac{e^{-ae^{W\left(-\frac{1}{2\cdot a}\right)}}}{\sqrt{-2\cdot W\left(-\frac{1}{2\cdot a}\right)}}.
\end{equation}
This score function is continuous and has a continuous first derivative which increases from $\pm \infty$ up to $\pm e^{-a e^{W\left(-\frac{1}{2 a}\right)}}$, after which the function quickly smooths out to a point of zero derivative at 0. Here the hyper-parameter $a$ determines how aggressively the scoring works, in both absolute amount and gradient. Increasing $a$ will shift the point of maximum gradient closer to zero leading to a more strict self-reinforcement and it also increases the maximum gradient and the rate of increase in the gradient, hence leading to a more aggressive training and self-reinforcement. In our experiments, we find that a Lambert cutoff between 2 and 5 works best. We report the results using a Lambert cutoff of 4.7, which we empirically found to work best in our experiments. 

\subsection{Conditioning-driven Diversity Loss}
We use the score computed by Eq.~\ref{eqn:10} as the quality term in the DPP kernel (Eq.~\ref{eqn:8}). This conditioning performance-based DPP loss models diversity and conditioning simultaneously and assigns a smaller loss value to samples that are diverse and meet conditioning requirements. Specifically, we construct the kernel matrix $L_{B}$ for a generated batch $B$ based on Eq.~\ref{eqn:8}:
\begin{equation}
\label{eqn:12}
L_{B}(i, j)=k\left(\mathbf{x}_{i}, \mathbf{x}_{j}\right)\left(q\left(\mathbf{x}_{i}\right) q\left(\mathbf{x}_{j}\right)\right)^{\gamma_{0}},
\end{equation}
where $x_{i},x_{j} \in B$, $q(x)$ is the quality function which determines the quality of the samples generated, and $k(x_{i},x_{j})$ is the similarity kernel between $x_{i} \text{and} x_{j}$. The exponent term $\gamma_{0}$ is introduced as a parameter to control the trade-off between the conditioning score and diversity. The conditioning performance-based DPP loss can be expressed as:
\begin{equation}
\mathcal{L}_{\mathrm{PcD}}(G)=-\frac{1}{|B|} \log \operatorname{det}\left(L_{B}\right)=-\frac{1}{|B|} \sum_{i=1}^{|B|} \log \lambda_{i},
\end{equation}
where $\lambda_{i}$ is the $i$-th eigenvalue of $L_{B}$. This loss will then be added to the overall loss of the generator during training. With this decomposition of the DPP kernel, we combine the objective of diversity and conditioning. In doing so, we not only promote good conditioning and design space coverage but also ensure consistent conditioning quality across the design space.

\subsection{Vicinity-based Loss}

 The vicinal loss defined in Eq.\ref{eqn:3} and Eq.\ref{eqn:4} relies on uniform sampling of labels from the training data. However, uniformly distributed labels are uncommon in real-world settings. This leads to potentially massive imbalances in the training with more common parts of the label space receiving more attention. To address this problem, we first randomly and uniformly sample a label between the minimum and maximum value (say 0 and 1 after normalization) and select the label present in the data which is closest to this random label. By doing so, we ensure even coverage of the conditioning space and avoid the bias towards more common labels during training.
 
To achieve conditional diversity (\ie, generate diverse designs given each condition/label), we introduce \textit{singular vicinal loss}. Rather than just selecting labels uniformly in the label space in each training step, we select only one label from the data and apply the vicinal loss only for this label's vicinity. The DPP loss would then encourage even coverage of the design space given each label as the condition. Intuitively, it will encourage designs which have similar performance but different forms (\eg{} shapes of airfoils), a much desirable outcome for exploration of design options.
Furthermore, the conditioning quality term of the DPP will promote accurate conditioning at all modes of the design space in this approach as the DPP loss will only be applied to the vicinity of one label. Therefore, by using the singular vicinal loss and the DPP loss, we simultaneously promote large design space coverage and accurate conditioning of performance labels across the design space.
The loss can be formulated by the same equations as Eq.\ref{eqn:3} and Eq.\ref{eqn:4} with the following applied to all $y_{j}$'s:
\begin{equation}
y_{j}=y_{s} \text { for all } j, y_{s} \in p\left(y_{\text {data }}\right),
\end{equation}
where $y_{s}$ is the singular labels selected for the given training step, which is selected by the process described at the beginning of this section. The overall PcDGAN loss can be written as:

\begin{equation}
\label{eqn:15}
{\mathcal{L}}^{PcDGAN}(D, G)={\mathcal{L}}^{\mathrm{sVDL}}(D) + {\mathcal{L}}^{\epsilon}(G) + \gamma_{1} {\mathcal{L}}^{PcD}(G),
\end{equation}
where ${\mathcal{L}}^{\mathrm{sVDL}}(D)$ refers to the singular vicinal loss. It can be either a singular hard vicinal discriminator loss (sHVDL) or a singular soft vicinal discriminator loss (sSVDL). ${\mathcal{L}}^{\epsilon}(G)$ is the vicinal loss term of the generator. ${\mathcal{L}}^{PcD}(G)$ is the conditioning performance-based DPP loss and the parameter $\gamma_{1}$ controls the weight this term.

\subsection{Condition Input Mechanism}
In this section, we propose a new way to effectively feed the condition label to the generator and discriminator in a GAN model.
The embedder model used by the ILI method in~\cite{CcGAN} has several issues: 1)~it is computationally expensive to train; 2)~it is difficult to map a condition to the features extracted from the estimator's last layer as the mapping can be one-to-many; and 3)~in many design applications, differentiable high-fidelity estimators for providing accurate label prediction are used, which means that there is no need for a neural network estimator and one cannot extract the features for training the embedder. 

\begin{figure}[h!]
\centering
\includegraphics[width=\columnwidth]{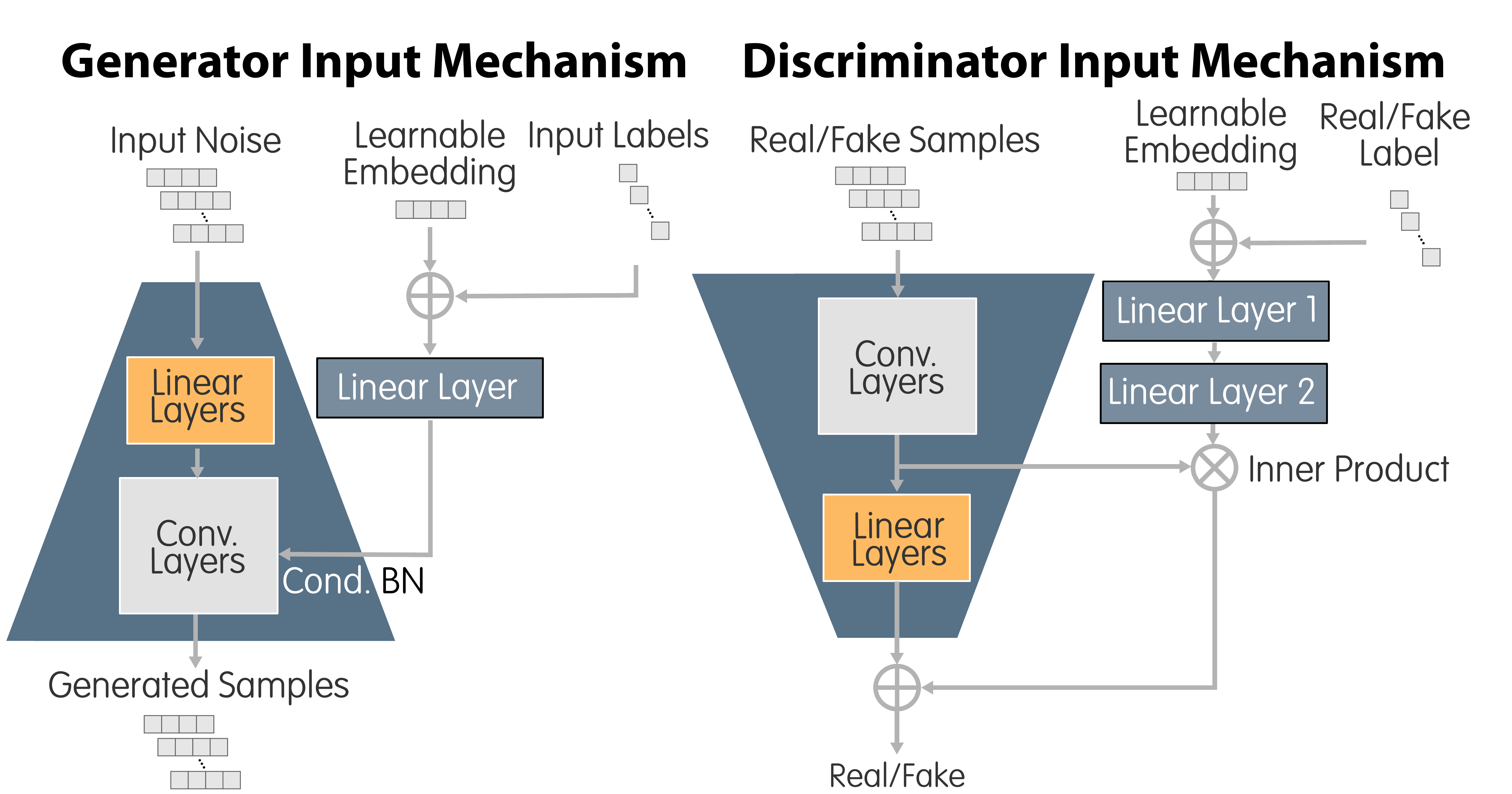}
\vskip 0.0in
\caption{Label input mechanisms used in PcDGAN for the generator (left) and discriminator (right)}
\label{fig:input}
\end{figure}

Instead of using an embedder, we start with a learnable embedding and add the condition labels to this embedding to get a vector $\boldsymbol{v}_0$. Then we transform $\boldsymbol{v}_0$ into $\boldsymbol{v}_1$ through a fully connected layer. We feed $\boldsymbol{v}_1$ to the generator's convolutional layers using conditional batch normalization~\cite{cbn} and to the discriminator by label projection\cite{projection}. The described mechanism is illustrated in Fig.~\ref{fig:input}. By proposing a simple learnable embedding and linear layer mechanism instead of using an embedder, we find that the conditioning improves significantly and the computational cost reduces. The architecture of this label input mechanism is shown in Fig.\ref{fig:input}.

\subsection{Evaluation Metrics}
We measure the performance of models by measuring how well they meet a given condition and how diverse the set of generated samples are. To measure the success of conditioning we evaluate two metrics. The first metric, named ``label error''(called label score in ~\cite{CcGAN}), is the mean absolute error between the input condition and predicted actual label of generated samples. This metric measures on average how close the sample label is to the target label and the closer a value is to zero, the better. Mean absolute error however, can be flawed in measuring conditioning performance due to its sensitivity to a few extreme ouliers, even if most samples are in the vicinity of the input label. To overcome this issue we also define another metric, named ``likelihood score'', which measures the likelihood of the input label under the generated samples' label distribution. To calculate this metric, we compute the Gaussian kernel density estimation (KDE) of the output samples' labels at the input condition (for every KDE we find the optimal bandwidth using a grid search). Unlike the label error, the likelihood score provides a more practical measure from a design perspective as it indicates how likely it is that the generated samples will meet a designer's requirements. The higher the likelihood score, the better.

To measure diversity in continuous space, we measure design coverage (\ie, diversity) achieved by a set of samples generated by each model using the log determinant of the similarity matrix of the samples (\ie, the DPP log-likelihood):
\begin{equation}
s_{\mathrm{div}}=\frac{1}{n} \sum_{i=0}^{n} \log \operatorname{det}\left(L_{S_{i}}\right),
\end{equation}
where $n$ is the number of times diversity is evaluated, $S_{i} \subseteq Y$ is a random subset of $Y$, the set of generated samples or training data. L is the similarity kernel. 
In the experiments, we set $|Y|=1000,\left|S_{i}\right|=10$, and $n=1000$. We evaluate the above metrics 10 times for 100 conditions ranging from 0.05 to 0.95. Each time we generate 1000 samples for each condition (100,000 samples each time, totaling a million samples for each model). To account for the stochasticity of the models, we evaluate all the metrics in 10 different training runs of each method.

\section{Experiments}
At this point, we have established our method and illustrated reasons and justifications for why we believe our methods will perform more accurate conditioning in continuous spaces. Here we perform experiments to establish validation for our claims and provide evidence of the performance of our model via two synthetic examples and a real-world inverse airfoil design example. For further details on the implementation please refer to our code and data for reproducing the experimental results\footnote{Code and data will be made open-source after paper acceptance}. The overall results of all experiments with respect to all metrics is presented in Table\ref{table:summary}.

\subsection{Synthetic Examples}
\begin{figure*}[h]
\centering
\includegraphics[width=2\columnwidth]{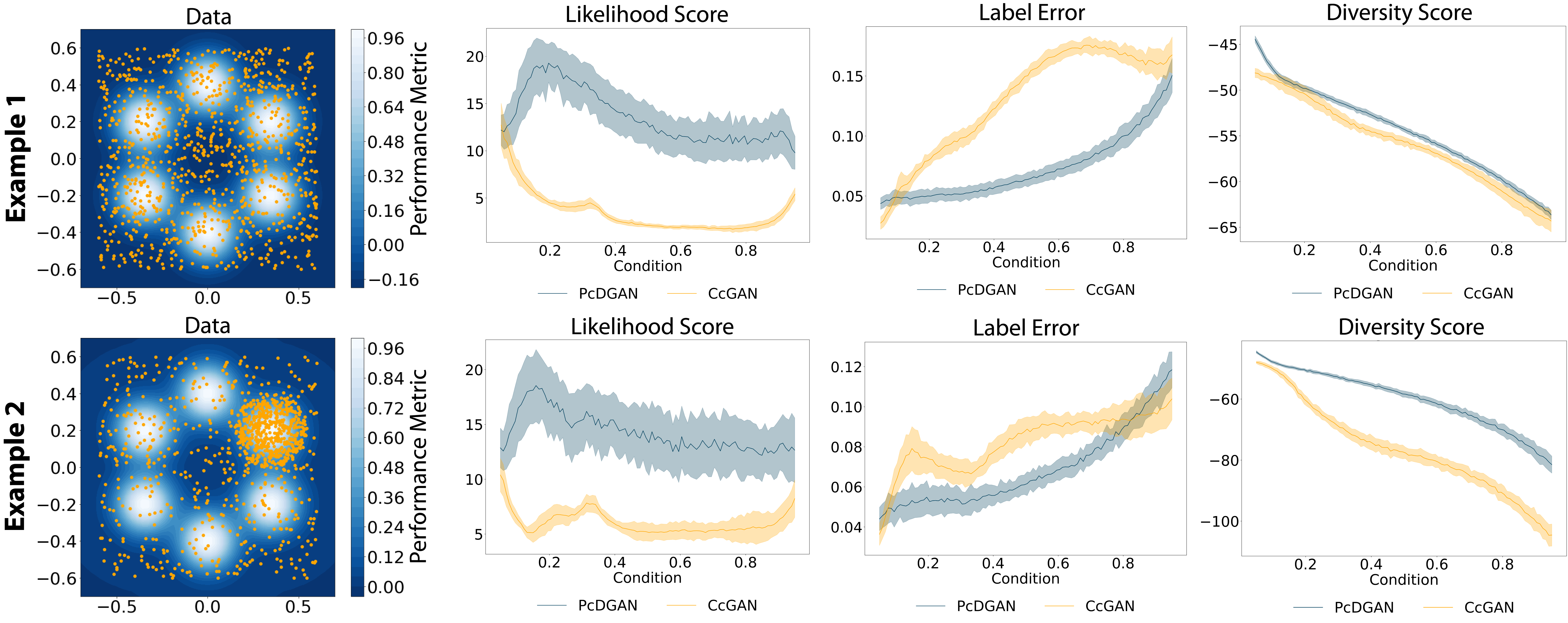}
\vskip 0.0in
\caption{Visualization of the synthetic examples. In the left-most images, orange dots indicate data-points. Only a sub-sample of 1,000 of the 10,000 data points is shown for clarity. The plots on the right indicate the mean and standard deviation of PcDGAN and CcGAN for all three metrics.}
\label{fig:simulated_examples}
\end{figure*}

To demonstrate PcDGAN's performance we create two simulated examples, shown in Fig.~\ref{fig:simulated_examples}, where the conditioning performance can be visually verified. In both examples we model the performance metrics as a density function of an un-normalized Gaussian mixture:
\begin{equation}
\label{eqn:18}
q(\mathbf{x})=\sum_{k=1}^{K} \exp \left(-\frac{\left(\mathbf{x}-\mu_{k}\right)^{T}\left(\mathbf{x}-\mu_{k}\right)}{2 \sigma^{2}}\right),
\end{equation}
where $\mu_{k}$ is the mode of the $k$-th mixture component and $\sigma$ is the standard deviation. The centers $\mu_{1}, \ldots, \mu_{K}$ are evenly spaced around a circle centered at the origin and with a radius of $0.4$. We set $K=6$ and $\sigma \approx 0.1$. Hence, there are six peaks in the conditioning space making the problem a multi-modal problem. For every input condition, there are points in all six modes and a method which can generate samples in all the modes is preferred. For the first example, the data is uniformly distributed within -0.6 and 0.6 of the origin in both directions hence simulating a case where the data is evenly covering the label space without any bias. For the second example a similar distribution exists for 50\% of the data, while the other 50\% of the data are located in a circle around one of the peaks (centered at $\mu_{2}$ with a radius of 0.2). The second example is used to demonstrate the case where the data is distributed unevenly in the design space, which may lead to mode collapse. In both examples, 10,000 data points are used for training.

\subsection{Inverse Design of Airfoils}
An airfoil is the cross-sectional shape of a wing and is commonly found in blades of propellers, rotors, and turbines. A common problem faced by practitioners, while designing an airfoil, is to find the right airfoil which meets a specific performance requirement. A practitioner also benefits from considering multiple design alternatives and picking the one which meets their requirement.
The most common way to measuring the performance of airfoils is the ratio of lift to drag, which we consider in this example. 
We show how PcDGAN can be used for the inverse design of airfoil shapes for any given performance target and it leads to a diverse set of design alternatives.
In this example, we use the UIUC airfoil database\footnote{\href{https://m-selig.ae.illinois.edu/ads/coord_database.html}{https://m-selig.ae.illinois.edu/ads/coord\_database.html}} as our data source. It provides the geometries of nearly 1,600 real-world airfoil designs. We pre-processed and augmented the dataset based on \cite{chen_fuge_2019} to generate a dataset of 38,802 airfoils, each of which is represented by 192 surface points (i.e. $x_{i}\in \mathbb{R}^{192 \times 2}$). We calculate the lift to drag ratio ($C_{L}/C_{D}$) for all airfoils and use it as a performance measure, on which the generative models are to be conditioned. 
To obtain an airfoil's performance, a computational fluid dynamics (CFD) simulation is typically required, for which we used the XFOIL software~\cite{xfoil}.
We scaled the performance scores between 0 and 1. For the estimator we trained a neural network-based surrogate model on all 38,802 airfoils to approximate $C_{L}/C_{D}$. The distribution of $C_{L}/C_{D}$ labels in the data is not uniform with a mean of 0.4507 and standard deviation of 0.1483. The data is very sparse at higher and lower values with only 0.18\% of the data above 0.9 and 0.5\% of the data below 0.1. The non-uniform distribution of labels makes it difficult for any data-driven method to generate samples in the sparse regions. We observe that PcDGAN's singular vicinal loss overcomes this issue.

\subsection{Model Configuration}
To demonstrate the performance of PcDGAN we compare it with the state of the art approach in continuous conditioning CcGAN.
For diversity we use an RBF kernel with a bandwidth of 1 to constructing $L_{B}$ in Eq.\ref{eqn:12}, \ie, $k(\mathbf{x}_{i}, \mathbf{x}_{j})=\exp (-0.5\|\mathbf{x}_{i}-\mathbf{x}_{j}\|^{2})$. This gives a value between 0 to 1, with a higher value for more similar designs. In the synthetic example we set $\gamma_{0}=3.0$ and $\gamma_{1}=0.5$ and in the airfoil example we set $\gamma_{0}=3.0$ and $\gamma_{1}=0.4$. Further, for the quality term in Eq.~\ref{eqn:12} in the airfoil example, we use a realistic conditioning quality\cite{padgan} which means that the conditioning quality will be multiplied by the discriminator output, i.e. $q(\mathbf{x})=D(\mathbf{x}) q^{\prime}(\mathbf{x})$, so as to promote GAN stability. For the LLETS, we set $a=4.7$ for all examples(this value was found to work best through empirical experimentation). Finally, in both methods, we use the soft vicinal loss (it was observed that the soft vicinal loss performed better in both methods across all examples) and pick $\kappa$ and $\sigma$ based on the rule of thumb method described by \cite{CcGAN}. In the Airfoil example we used a residual neural network (ResNet)\cite{ResNet} trained on the dataset as the estimator model and a BézierGAN\cite{BezierGAN} to generate airfoils. In this example we refer to the continuously conditioned BézierGAN as `CcGAN' and the BézierGAN with $\mathcal{L}_{PcD}$ as `PcDGAN'. In the airfoil example we train an embedder model exactly as implemented in \cite{CcGAN} for the CcGAN model using the ResNET estimator. For the synthetic examples the exact equation (Eq.\ref{eqn:18}) was used as the estimator for PcDGAN.

\section{Results and Discussion}
In this section we report the results of applying PcDGAN to two synthetic and one real-world example.
\subsection{Synthetic Example}
\begin{figure}[H]
\centering
\includegraphics[width=\columnwidth]{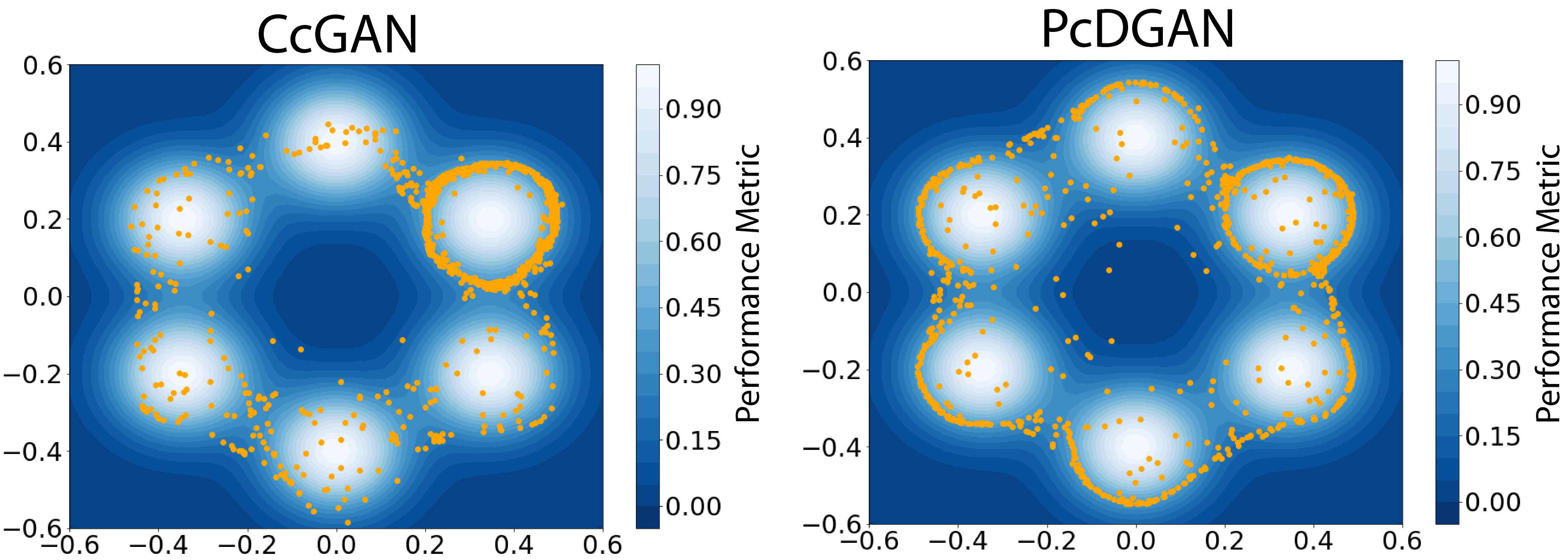}
\vskip 0.0in
\caption{Comparison of output distribution of 1000 generated data points, shown by orange dots, from CcGAN and PcDGAN for input condition of 0.4 in Example 2.}
\label{fig:visual_diversity}
\end{figure}

The results of the experiments on both examples are presented in Fig.\ref{fig:simulated_examples} which shows the statistics of each model in all three performance metrics across ten runs. Both examples show that PcDGAN outperforms CcGAN in all three metrics.
In conditioning we observe that in the first example CcGAN has higher label error and worse likelihood performance compared to example 2, while PcDGAN's performance remains consistent and better than CcGAN in both examples. This difference in conditioning performance can be explained by the difference in diversity. We observe that despite PcDGAN performing better in diversity in the first example the difference between the diversity of the models is insignificant, meaning that both models cover the design space well with PcDGAN performing slightly better, which is expected as the dataset covers the design space uniformly. In the second examples however we observe a large gap in the diversity performance of CcGAN and PcDGAN, with PcDGAN performing similar to the first example but CcGAN suffering in producing diverse samples. This indicates that CcGAN produces samples mostly in the dominant region, which means that the model will only have to condition well on one mode, which is a much easier task than doing so across all modes. This is the underlying cause of the improvement in CcGANs performance. This is all while PcDGAN performs consistently across both examples in both conditioning and diversity. This difference in diversity can be visually verified in these 2D examples. In Fig.\ref{fig:visual_diversity} the samples generated by both models conditioned on 0.4 is visualized(the same patterns are observed across all conditions). Ideally, any good design generation model should cover all modes of the data despite the imbalance in the dataset. In Fig.\ref{fig:visual_diversity} we observe that PcDGAN has produced samples in all modes while also maintaining very good conditioning across all modes, however we see that CcGAN not only fails to cover all modes as evenly as PcDGAN but it also fails to produce well conditioned samples in less common modes when it does produce samples in those regions. PcDGAN's performance is due to the functionality of the conditioning performance-based DPP loss which not only promotes good coverage but also promotes high quality conditioning across all modes.

\subsection{Inverse Airfoil Design}
\begin{figure*}[ht]
\centering
\includegraphics[width=1.7\columnwidth]{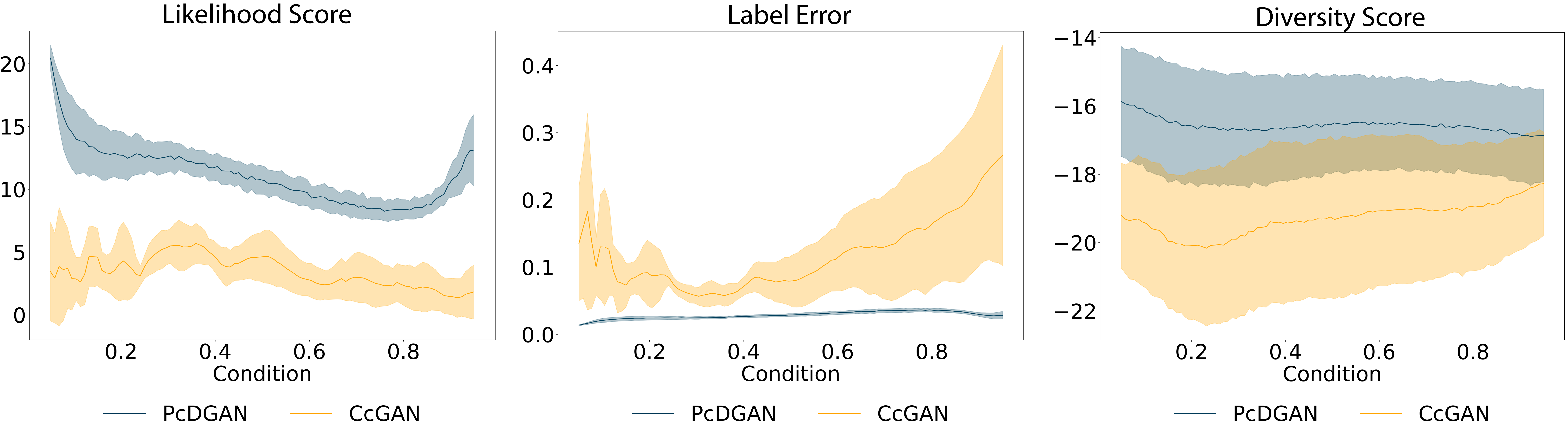}
\vskip 0.0in
\caption{Comparison of performance between CcGAN and PcDGAN for different conditions in the airfoil example. The graph indicates the mean and standard deviation. PcDGAN shows better performance and lower variation across runs for all metrics.}
\label{fig:airfoil_results}
\end{figure*}
\begin{figure*}[h!]
\centering
\includegraphics[width=1.8\columnwidth]{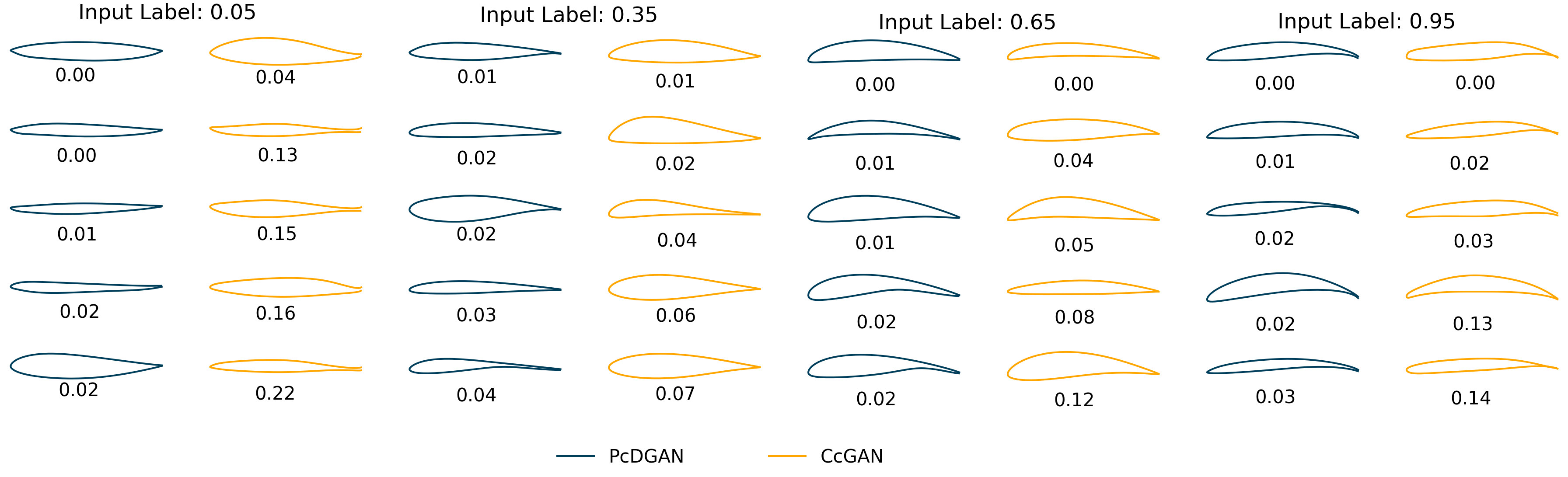}
\vskip -0.1in
\caption{Airfoil shapes generated by each model for different conditions, with their label error shown below every airfoil.}
\label{fig:airfoil_samples}
\end{figure*}

\begin{figure*}[h!]
\centering
\includegraphics[width=1.8\columnwidth]{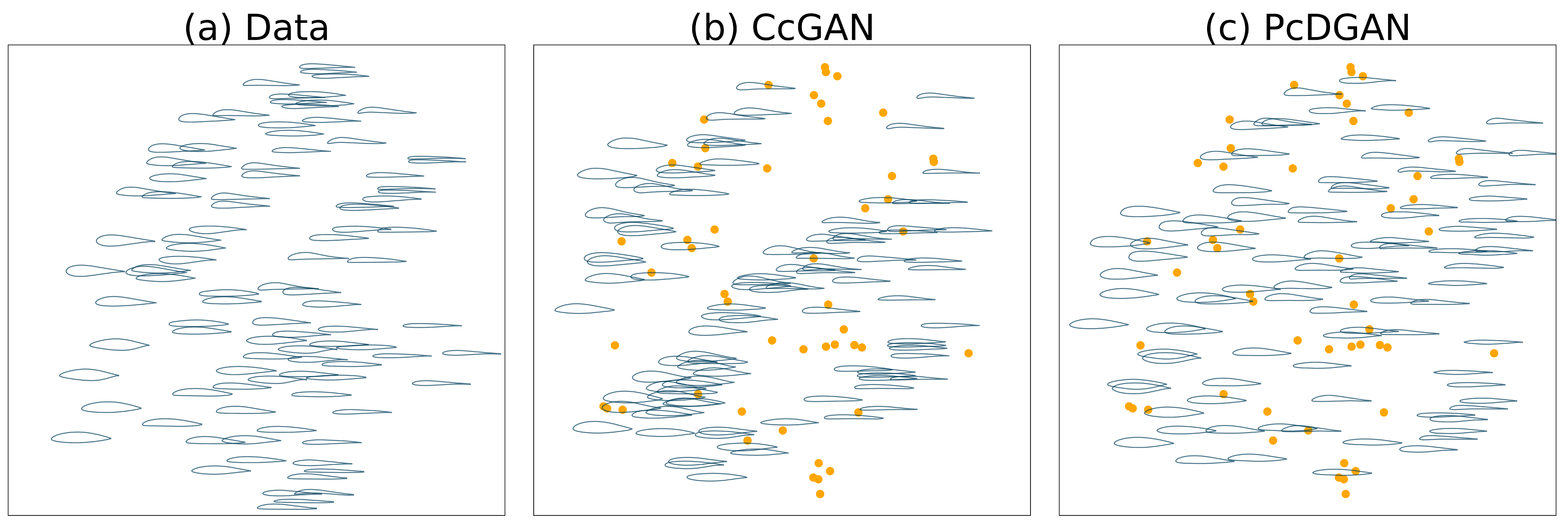}
\vskip -0.1in
\caption{We show airfoils corresponding to samples from a) data, b) CcGAN and c) PcDGAN, all conditioned on 0.3 with with the predicted labels within $0.3\pm0.025$. All samples are mapped to the same 2D space using t-SNE. Orange dots in plots (b) and (c) indicate embeddings of the real data. This shows that PcDGAN samples are more diverse and cover a larger design space.}
\label{fig:airfoil_tsne}
\end{figure*}
\begin{table*}[h!]
  \caption{Experimental results (mean and standard deviation)}
  \vskip -0.1in
  \label{table:summary}
  \label{tab:freq}
  \begin{tabular}{ccccc}
    \toprule
    & Model & Likelihood Score & Label Error & Diversity Score\\
    \midrule
    \multirow{2}{*}{\textbf{Ex.~1}}& CcGAN & 3.494$\pm$2.307& 0.1296$\pm$0.0452& -55.817$\pm$4.637\\
    & PcDGAN &\textbf{13.604$\pm$3.628}&\textbf{0.0725$\pm$0.0288}& \textbf{-54.537$\pm$4.728}\\
    
    \multirow{2}{*}{\textbf{Ex.~2}}& CcGAN & 6.043$\pm$ 1.383&0.082$\pm$0.012& -76.412$\pm$14.896\\
    & PcDGAN &\textbf{14.280$\pm$3.213}&\textbf{0.068$\pm$0.022}&\textbf{-59.938$\pm$9.468}\\
    
    \multirow{2}{*}{\textbf{Airfoil}}& CcGAN & 3.429$\pm$2.240& 0.119$\pm$0.026& -19.297$\pm$2.140\\
    & PcDGAN & \textbf{11.181$\pm$2.686}&\textbf{0.0284$\pm$0.004}& \textbf{-16.567$\pm$1.509}\\
    
  \bottomrule
\end{tabular}
\end{table*}
We observe that across all metrics PcDGAN performs well and outperforms CcGAN (Fig.\ref{fig:airfoil_results}) with an average performance gain of 7.75 on the likelihood score from 3.429 (CcGAN) to 11.181 (PcDGAN), an average reduction of 0.09 in label error from 0.119 (CcGAN) to 0.0284 (PcDGAN) and an average gain of 2.73 in the diversity score from -19.297 (CcGAN) to -16.567 (PcDGAN) (Table\ref{table:summary}). The low label errors and high likelihood scores across all conditions shows that the performance of the generated samples is closely met by PcDGAN. For all conditions, the average diversity of samples is also higher in PcDGAN which indicates that PcDGAN covers the design space better and consistently across all conditions. It is important to note that in PcDGAN, the variation of metrics across conditioning space for different runs is low compared to CcGAN, which has high variations.
To further demonstrate the performance of each approach, we use t-Distributed Stochastic Neighbor Embedding (t-SNE) to map airfoils onto a two dimensional space. The results of this are shown in Fig.\ref{fig:airfoil_tsne}. To compute t-SNE we use 1000 samples from the data with labels in the vicinity of the input label 0.3 (labels within $0.3\pm0.025$), then we generate 1000 samples using PcDGAN and another 1000 samples using CcGAN both conditioned on the input label of 0.3 and with predicted labels within $0.3\pm0.025$ to visualize the output distributions of useful (\ie, near target performance) samples generated by each model. We then plot the results of 50 samples from each of the data, PcDGAN, and CcGAN in Fig.\ref{fig:airfoil_tsne}. Further we display these airfoils and their location in the t-SNE plot on Fig.\ref{fig:airfoil_tsne}(a)-(c). we observe that CcGAN has failed to cover the entire sample space evenly with gaps existing in the data that are not covered by CcGAN. This is while PcDGAN has been able to fill in some of the gaps in the data (\ie, interpolate) with useful samples that have labels much closer to the input label condition. Finally, in Fig.\ref{fig:airfoil_samples} we plot the geometries of some of the generated samples at different conditions using both approaches and display their respective label errors below them. It is observed that PcDGAN generated samples have lower label errors across all labels. Furthermore, it is observed that the quality of airfoils generated by CcGAN at low input labels(Left most column in Fig.\ref{fig:airfoil_samples}) are far lower than PcDGAN with some airfoils having open trailing edges and unrealistic shapes. This is caused by the fact that the number of samples with labels in that vicinity are lower, which means that CcGAN will not be trained on those labels as often as other labels, which has results in lower quality and unrealistic samples being generated for less common labels.

\section{Conclusion and Future Work}
In this work, we introduced a novel method for conditioning GANs in continuous spaces while promoting diversity in the generated samples and allowing for the direct use of any differentiable estimator (exact or regression estimator) to improve conditioning. We achieved this by introducing a novel method for measuring the quality of conditioning called ``LLETS" with self-reinforcing properties. We integrated LLETS with a DPP based diversity loss function to promote diversity and high-quality conditioning simultaneously and name the resultant GAN model as ``PcDGAN''. 
In PcDGAN, we also introduced the singular vicinal loss to overcome data imbalance and sparsity challenges and reduce performance variation. 
We show through both synthetic and real-world experiments that PcDGAN out-performs the current state of the art in continuous conditioning CcGAN in both producing diverse and well-conditioned samples. Furthermore, we show that even when few samples exist in some parts of the conditioning space (less than 0.2\%), PcDGAN can produce samples with a high likelihood of meeting input conditions while prior methods simply fail to do so.

PcDGAN was developed with design applications in mind, where engineering design space exploration requires models that can generate samples covering the entire design space while meeting certain requirements(\ie, conditions). We show that PcDGAN is capable of accomplishing both objectives of design space exploration. It is capable of generating new designs that are highly likely to meet design requirements while covering the entire design space and sometimes expanding it to fill in the gaps in the data distribution. 
These properties make PcDGAN an ideal approach in data-driven design synthesis and design space exploration, which can be used as a valuable tool for inspiring new engineering designs. 
Despite being design-centric, our method can be generalized to other domains or more complex data-driven synthesis tasks. One example could be 3D design synthesis or CAD generation, where PcDGAN can be trained on a 3D dataset to produce CAD models that can be conditioned to meet physical and mechanical constraints and performance requirements. Other applications include image generation, metamaterial design or parametric design.

Finally, even though we demonstrate our methods in GANs, the DPP based loss or the LLETS alone can be used in any other data-driven synthesis method such as VAEs. Furthermore, our method can be expanded to multi-objective problems with the same LLETS quality applied across a vector of conditions rather than a single condition. Future work will focus on adapting the singular vicinal loss to higher dimensional conditions when the data sparsity problem is more severe and explore PcDGAN's integration with objective reinforced learning for problems where training a differentiable performance estimator is not feasible. 

\bibliographystyle{ACM-Reference-Format}
\bibliography{acmart}

\pagebreak
\pagebreak
\appendix

\section{Implementation and Model architecture Details}
In this section we will discuss some of the detailed explanation of our implementation and training parameters for each of the experimental examples.

\subsection{Airfoil Example}
\subsubsection{ResNet Estimator}
\begin{figure}[h!]
\centering
\includegraphics[width=\columnwidth]{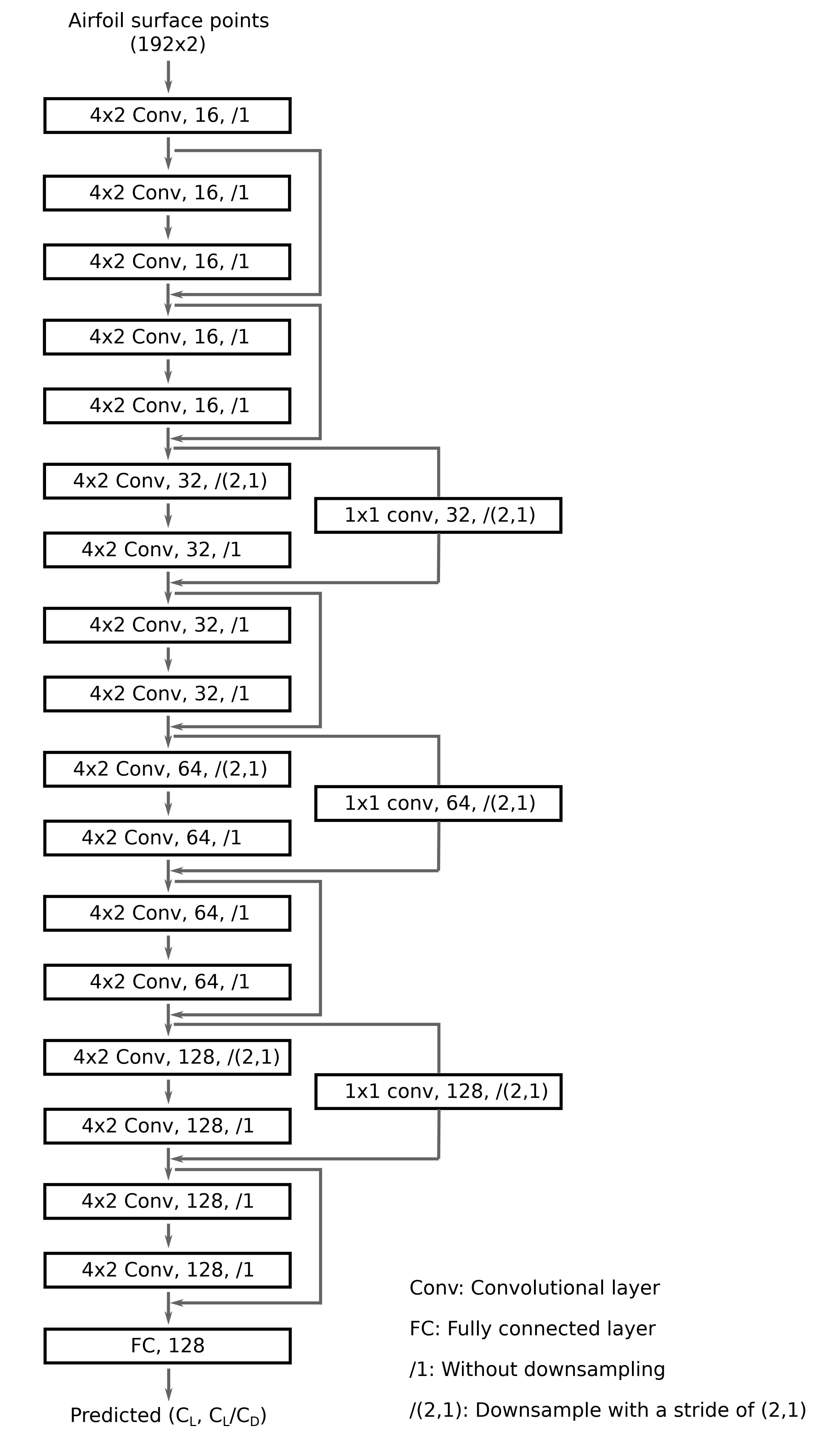}
\vskip 0.0in
\caption{ResNet Architecture of the estimator}
\label{fig:surrogate}
\end{figure}

As discussed in the body of the paper we train a ResNet model as a surrogate estimator in place of an exact estimator to predict the lift to drag ratio of any given airfoil. This model is then used during the training of PcDGAN. Further this model is what we use to train the embedder model of CcGAN. The architecture of this model is presented in Fig.\ref{fig:surrogate}. Given the fact that the data is very imbalanced if the surrogate model is directly trained on the data it will form a bias towards more common labels which would negatively effect both CcGAN and PcDGAN's performance. Given this during the training of the estimator we select balanced minibatches, that is to say we sample data such that the labels cover the label space uniformly (Sample random uniform numbers between minimum and maximum label and pick the sample with the label closest to the random number). A similar approach is taken in training the embedder of the CcGAN. We observe that because we use the estimator to report conditioning results if the estimator is not balanced real-life performance of airfoils will not be as conditioned as well in neither CcGAN nor PcDGAN, however the trend of the overall performance data based on the estimator which is reported in the paper does not change in comparing the two models. Our code uses XFOIL to verify that the trends in conditioning of all methods do translate to exact numerical simulations of the airfoils (refer to section A.3 for code). Finally, for training both the estimator and embedder we use Adam optimizer with a base learning rate of $10^{-4}$ which decays with a multiplier of 0.46 every 2500 of training steps (staircase decay) and we train both models for 10,000 steps with a batch size of 256 and apply ealry stopping based on the performance of the models on a smaller subset of the data used for validation and testing. The architecture of the CcGAN embedder model is identical to the CcGAN author's approach described in \cite{CcGAN}.
\subsubsection{BézierGAN Training Details}
To generate airfoils we use the BézierGAN architecture which has been used successfully to generate realistic curves and specifically airfoils. Our implementation is identical to the authors implementation \cite{BezierGAN} with the only difference being in our added label input mechanism. For the generator we replace all batch normalization layers with conditional batch normalization which is conditioned on the embedder output in CcGAN and on the label embedding in PcDGAN. To obtain label embedding in PcDGAN each of the generator and discriminator have separate label embedding mechanism as described in Fig.\ref{fig:input}. To keep the size of the model the same as in CcGAN, the linear layer we use to go from the learnable embedding to the input of the conditional batch normalization has an output size of 128. The learnable embedding itself has a size of 10, which makes this mechanism much smaller than the 7 layers of size 128 used in the CcGAN ILI embedder. For the discriminator we use label projection for both CcGAN and PcDGAN as described by Fig.\ref{fig:input}. Note that before the inner product in the label projection we apply spectral normalization\cite{Spectral} for both CcGAN's and PcDGAN's linear layer output which maps embeddings to a vector of the correct size suitable for applying the inner product(liear layer 2 in Fig.\ref{fig:input}). In training we use the Adam optimizer with a base learning rate of $10^{-4}$ which decays with a multiplier of 0.8 every 2,000 steps. We train both models for 20,000 steps with a batch size of 32. Furthermore, it is important to note that the authors of BézierGAN \cite{BezierGAN} use a separated discriminator training(\ie, the discriminator is trained in two step, one on the fake samples and one on the real samples), however this approach rendered CcGAN entirely non-functional, therefore, we used mixed training for CcGAN in this example while using the original implementation in PcDGAN as it did not have any negative results on PcDGAN.

Finally, it is important to mention that for the training of PcDGAN in the airfoil example an escalating schedule for $\gamma_{1}$ (DPP loss weight) is implemented. PcDGAN is more likely to generate unrealistic designs in early stages of training. Thus, $\gamma_{1}$ is initialized at 0 and increases during training, so that PcDGAN focuses on learning realistic designs at the early stage, and takes conditioning and diversity into consideration later. The schedule is set as:
$$
\gamma_{1}=\gamma_{1}^{\prime}\left(\frac{t}{T}\right)^{p}
$$
where $\gamma_{1}^{\prime}$ is the value of $\gamma_{1}$ at the end of training, $t$ is the current training step, $T$ is the total number of training steps, and $p$ is a factor controlling the steepness of the escalation which we set to 5.0 in out training. This is not necessary for the synthetic examples.
\subsection{Synthetic Example}
\begin{figure}[h!]
\centering
\includegraphics[width=\columnwidth]{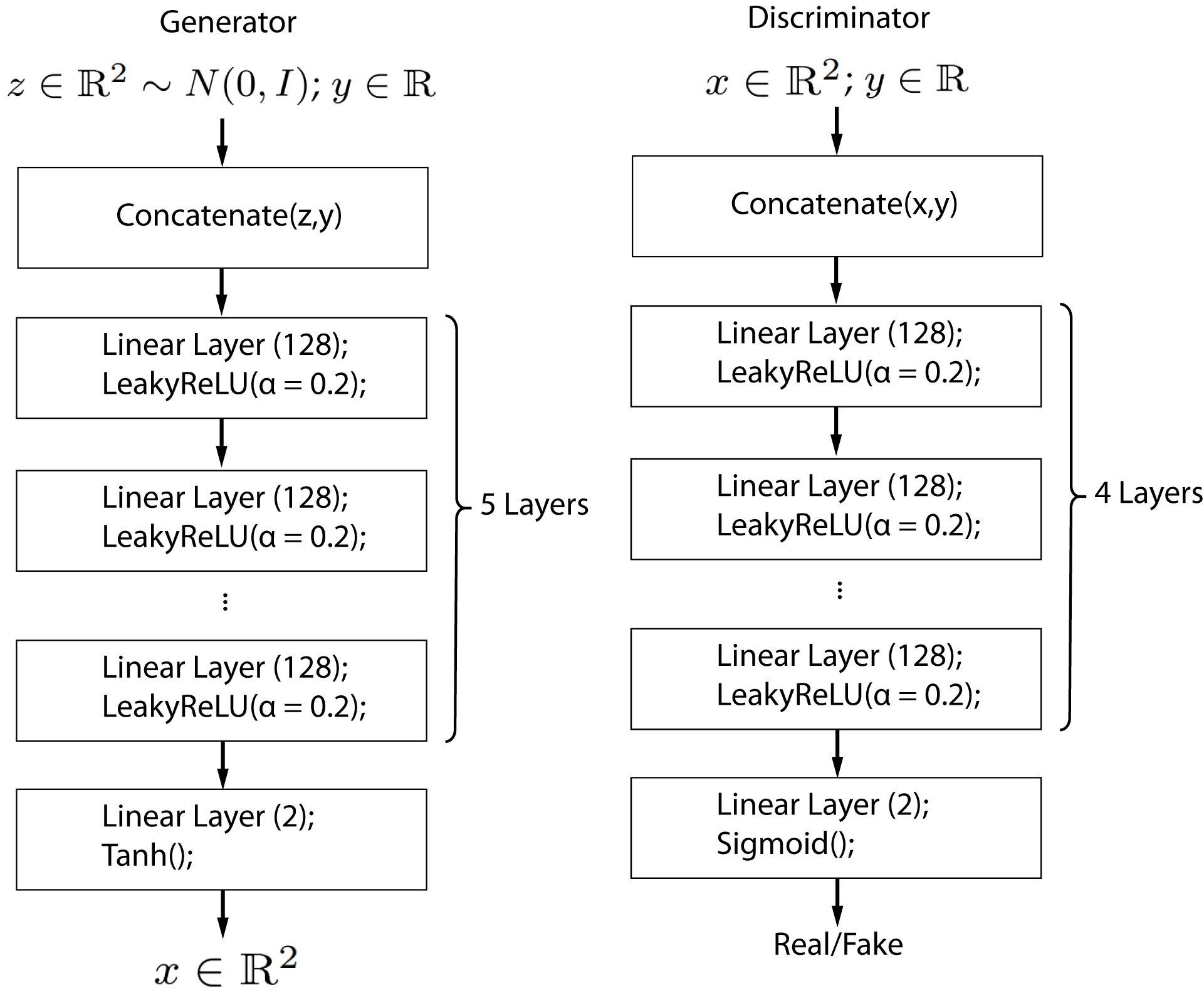}
\vskip 0.0in
\caption{Architecture of model used in the synthetic examples}
\label{fig:synth_model}
\end{figure}

In the synthetic example the models used have very simple architectures described in Fig.\ref{fig:synth_model}.

For the synthetic example we simply concatenate the input label and the rest of the inputs for both the discriminator and the generator. In these simple examples we the need for a complex label input mechanism does not exists and the authors of CcGAN\cite{CcGAN} also purport to this in their implementation. We observed no significant improvement in the performance of either method using other label input mechanisms similar to the mechanism used in the convolutional neural network. In fact we observed the performance of both models decline when more complex mechanisms were used. 

Furthermore, since the exact equation of performance is known in synthetics examples there is no need to train a neural network based estimator for PcDGAN and the exact equation is used as the estimator. This is similar to a case where an exact estimator exists. Moreover, since the label input mechanism is reduced to simply concatenating at the input CcGAN also does not need an embedder model, which means that we did not train any other models besides the GANs in this example.

For training we use the Adam optimizer with a base learning rate of $10^{-4}$ which decays with a multiplier of 0.8 every 5,000 steps. We train both models for 50,000 steps with a batch size of 32.

\subsection{Code and GitHub Repository}
The full implementation of the code used for this paper along with the airfoil dataset will be made available and open source upon the acceptance of this paper. The Github repository can be found under the url below:

\href{https://github.com/pcdgan/PcDGAN}{https://github.com/pcdgan/PcDGAN}

With regards to referenced architecture and methodology and algorithms, the code replicates these methodology as closely as possible to the authors' best knowledge and without any alterations unless mentioned in this section(such as the mixed discriminator training in BézierGAN implemented to improve CcGAN Results).
\end{document}